\documentclass[10pt, conference, compsocconf]{IEEEtran}
\usepackage{amssymb}
\usepackage{pifont}
\newcommand{\xmark}{\ding{55}}%
\usepackage{booktabs}
\usepackage[linesnumbered]{algorithm2e}
\usepackage{subfigure}
\usepackage{graphicx}
\usepackage[table]{xcolor}
\usepackage{tabularx}
\usepackage{multirow}
\usepackage{amsmath}
\usepackage{hyperref}
\usepackage{floatrow}
\usepackage{multicol}
    \newcolumntype{L}{>{\raggedright\arraybackslash}X}

\usepackage{tikz}

\newcommand{\DrawPercentageBar}[1]{%
  \begin{tikzpicture}
    \fill[color=green]   (0.0 , 0.0) rectangle (#1*3ex , 1.5ex );
    \fill[color=gray] (#1*3ex  , 0.0) rectangle (3.0ex, 1.5ex);
  \end{tikzpicture}%
}

\ifCLASSINFOpdf
\else
\fi
\begin{document}
%

\title{Open Domain Suggestion Mining Leveraging Fine-Grained Analysis}


\author{\IEEEauthorblockN{Shreya Singal*, Tanishq Goel*, Shivang Chopra*, Sonika Dahiya}
\IEEEauthorblockA{Delhi Technological University, Delhi \\
Email: \{shreyashivam20, tanishqgoel, shivangchopra11, sonika.dahiya11\}@gmail.com}

}


%


\maketitle

\begin{abstract}
Suggestion mining tasks are often semantically complex and lack sophisticated methodologies that can be applied to real-world data. The presence of suggestions across a large diversity of domains and the absence of large labelled and balanced datasets render this task particularly challenging to deal with. In an attempt to overcome these challenges, we propose a two-tier pipeline that leverages Discourse Marker based oversampling and fine-grained suggestion mining techniques to retrieve suggestions from online forums. Through extensive comparison on a real-world open-domain suggestion dataset, we demonstrate how the oversampling technique combined with transformer based fine-grained analysis can beat the state of the art. Additionally, we perform extensive qualitative and qualitative analysis to give construct validity to our proposed pipeline. Finally, we discuss the practical, computational and reproducibility aspects of the deployment of our pipeline across the web.
\end{abstract}


%
\IEEEpeerreviewmaketitle

\section{Introduction}
\subsection{Context and Original Scope}

The rising popularity of online review forums like Yelp\footnote{\href{https://www.yelp.com}{}}, and Tripadvisor\footnote{\href{https://www.tripadvisor.in}{}} has spawned a popular line of study involving suggestion mining. Suggestion Mining can be defined as the classification of reviews as Suggestion or Non-Suggestion which helps firms enhance their services according to the customers' needs. Manually going through the large number of reviews and filtering out the relevant suggestions is a very cumbersome process. Therefore companies now emphasize on trying to find robust automated suggestion mining mechanisms.

Apart from general suggestions, decision makers in big corporates aim to extract focused suggestions to improve their brands or organizational practices \cite{jijkoun-etal-2010-mining}. This in turn leads to a much less studied sub-task of open-domain suggestion mining \cite{odsm}. Unlike most of the work done currently in in-domain suggestion mining, open-domain mining involves a multi-tier classification wherein the model not only segregates the suggestions from the non-suggestions, but also predicts the specific domain to which the review belongs. 

\begin{figure}
    \includegraphics[scale=0.38]{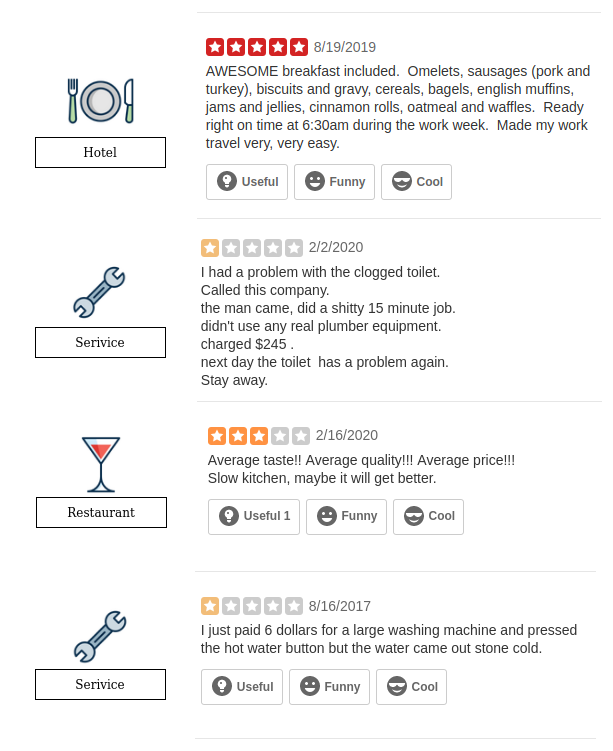}
    \caption{Manually provided multi-domain reviews on Yelp}
    \label{fig:reviews}
\end{figure}{}

A few examples of the kinds of reviews to be analysed are shown in Figure \ref{fig:reviews}. As clear from the examples, the reviews might or might not contain valuable suggestions. Many times reviews contain positive feedback of the customers' experiences which are not very valuable to organizations in terms of their utlity. The first example in the figure, although pretty detailed, doesn't add any valuable information which the hotel can make use of to improve its services. On the other hand, the other reviews are fairly clear in the problems with the respective products or services and hence give very clear indication to the respective firms of the improvements needed.

\subsection{Challenges}
Analysis of the current work done in the domain reveal several challenges to the correct classification of a review. The foremost being the sparse nature of suggestions in such a large pool of reviews. In-depth study of the various review platforms reveal a huge imbalance in the number of suggestions as compared to the number of non-suggestions. This imbalance induces unintended bias in the final classification pipeline which need to be eliminated. Apart from that, another significant challenge is the very long nature of reviews. In most of the cases, the length of a review is very big as compared to the actual suggestion presented in the review. Filtering out the required text from the review poses a hurdle to the correct classification of the respective reviews. The multi-domain nature of the data poses yet another challenge to the baseline classifiers. Correctly classifying suggestions and the domain they belong to proves to be a herculean task for those classifiers.

\subsection{Motivation}
The goal of our experiments is to gain insights into the way people formulate opinions regarding certain products or services and the differentiating dynamics of the cross-domain suggestions. Recent advancement in the field of Natural Language Processing and its application in the domain of Suggestion Mining has allowed companies to leverage customer reviews in order to enhance their products and services. The long process of manually segregating suggestions from the non-suggestions is financially taxing for the parent firms. An automated mechanism for open-domain suggestion mining significantly cuts down on these time and cost overheads. Furthermore, in most scenarios, a clear demarcation on the domain of the review is not present. A domain-independent system can therefore be effectively leveraged in such cases.

\subsection{Contribution}
The main contributions of our work are as follows:
\begin{itemize}
    \item Propose a novel two-phase transformer-based architecture for open-domain suggestion mining using fine-grain analysis.
    \item Leverage Attention mechanism to achieve state-of-the-art performance in suggestion mining.
    \item Employ discourse marker based oversampling to handle the imbalance nature of the data.
    \item Use an Adapter-based transfer learning mechanism in conjugation with Transformer to improve training efficiency and performance.
    \item Perform and in-depth qualitative and quantitative analysis to study the practical, reproducibility and deployment aspects of our proposed pipeline.
\end{itemize}

\begin{table}[ht]
    \centering
    \begin{tabular}{|c|c|c|c|c|}
        \hline
         \bf Baseline & {\bf ID} & {\bf OD} & {\bf DB} & \bf AM   \\
         \hline
         Ramanand et al. \cite{ramanand-etal-2010-wishful} & \checkmark & \xmark & \xmark & \xmark \\
         \hline
         Brun et al. \cite{Brun2013SuggestionMD} & \checkmark & \xmark & \xmark & \xmark \\
        \hline
        Wicaksono et al. \cite{hmm} & \checkmark & \xmark & \xmark & \xmark \\
        \hline
        Negi et al. (a) \cite{negi-buitelaar-2015-towards} & \checkmark & \xmark & \xmark & \xmark \\
        \hline
        Negi et al. (b) \cite{odsm} & \checkmark & \checkmark & \xmark & \xmark \\
        \hline
        Negi et al. (c) \cite{distant} & \checkmark & \checkmark & \xmark & \xmark \\
        \hline
         Jain et al. \cite{maitree} & \checkmark & \checkmark & \checkmark & \xmark \\
         \hline
         Ours & \checkmark & \checkmark & \checkmark & \checkmark \\
         \hline
    \end{tabular}
    \caption{Relative comparison of various baselines. \textbf{ID}: In-Domain Suggestion Mining \enspace \textbf{OD}: Open-Domain Suggestion Mining \textbf{DB}: Data Balancing \textbf{AM}: Attention Modelling}
    \label{tab:base}
\end{table}

\section{Related Work}
\subsection{Suggestion Mining}
The initial attempts at suggestion mining involved manual feature and linguistic rule modelling. Ramanand et al. \cite{ramanand-etal-2010-wishful} proposed lexicon-based rules to extract wishful comments from within suggestions. Further, Brun et al. \cite{Brun2013SuggestionMD} used semantic and morphological parsers to mine patterns and model linguistic rules to perform the task of suggestion mining. Following these, the focus shifted towards statistical machine learning approaches. Supervised classification was performed using Hidden Markov Models and Conditional Random Fields \cite{hmm}, Factorization Machines \cite{fm} and Support Vector Machines \cite{negi-buitelaar-2015-towards} to capture statistical-linguistic traits in the data. The use of neural networks for suggestion mining was first proposed by Negi et al. (a) \cite{negi-buitelaar-2015-towards}. Very recently, the SemEval challenge \cite{negi-etal-2019-semeval} saw several researchers perform the in-domain suggestion mining tasks using approaches like LSTM, BiLSTM, BERT \cite{bert}, ULMfit \cite{ulmfit}. All these approaches were however constrained to a single domain. 

\subsection{Open Domain suggestion Mining}
Negi et al. (b) \cite{odsm} first introduced the concept of open-domain suggestion mining. They used neural networks and pre-trained word embeddings to perform cross-domain training and also released the Open-Domain Suggestion Mining dataset which is subsequently used in our experiments. Building on this work, Negi et al. (c) \cite{distant} used an LSTM-based architecture to induce distant supervision in the Open-Domain Suggestion Mining task. This work was further extended by \cite{maitree} where they perform suggestion mining as a multi-task approach to classify the nature as well as the domain of the given reviews. Our work follows the footsteps of \cite{maitree} in modelling the nature of our downstream task.

\subsection{Imbalanced Classification}
Data imbalance has been a pressing issue in Natural Language Processing tasks for a long time, such as in Ghosh Chowdhury et al. \cite{ghosh-chowdhury-etal-2019-youtoo}. This a particularly grave issue in the case of suggestion mining due to the sparsity of the number of suggestions in the reviews. Imbalance data can be handles using two types of techniques:
\begin{itemize}
    \item \textbf{Random Undersampling}: Random undersampling involves randomly removing instances of the majority class to make the class-distribution even. However, this approach leads to a huge loss of data and is therefore not usually preferred.
    \item \textbf{Random Oversampling}: Traditionally, oversampling techniques like SMOTE (Synthetic Minority Over-Sampling Technique) \cite{smote} have been used to curb the problem of data imbalance. Further extensions of SMOTE like MSMOTE (Modified Synthetic Minority Over-Sampling Technique) \cite{msmote} and MWMOTE (Majority Weighted Minority Oversampling Technique) \cite{mwmote} have also been proposed. 
\end{itemize}

In the case of random oversampling, all the above approaches work in the euclidean space and hence, interpretability is compromised which is not acceptable in Natural Language Analysis tasks. This problem was effectively tackled by Jain et al. \cite{maitree} where they propose LMOTE (Language Model-based Oversampling Technique) which leverages a Hidden Markov Model to oversample minority classes. Following this, Zhang et al. \cite{discourse} used a discourse marker-based technique to perform oversampling on the Sentiment Analysis of imbalanced datasets to yield significant better performance in an interpretable fashion.

\section{Preliminaries}
\subsection{Problem Formulation and Notations}
Let $\mathcal{R} = \{ r_1, r_2,..., r_n \}$ be the set of $n$ reviews that belong to $d$ domains $\mathcal{D} = \{ d_1, d_2,..., d_d \}$. The purpose of our experiments is to perform multi-level classifcation to predict the nature $n_{r_i} \in \{ `suggestion', `non-suggestion' \}$ at the first level and then the domain $d_j$ of each review $r_i$ such that $n_{r_i} \in \{ `suggestion' \}$. 

We use $V$ to denote the vocabulary of English tokens present in our dataset. A $d$-dimensional word embedding $e_i$ is associated with each token $\mathbf{v_i} \in V$ such that $e_i \in R^d$. The set of all non zero embedding vectors is represented as $E$. Vectors are displayed in boldface. The embedding vector $e_i$ for each token $v_i$ can be mathematically defined as:
\begin{equation}
e_i = Word2Vec(v_i)
\end{equation}

\begin{figure}[t]
    \centering
    \includegraphics[width=0.9\linewidth]{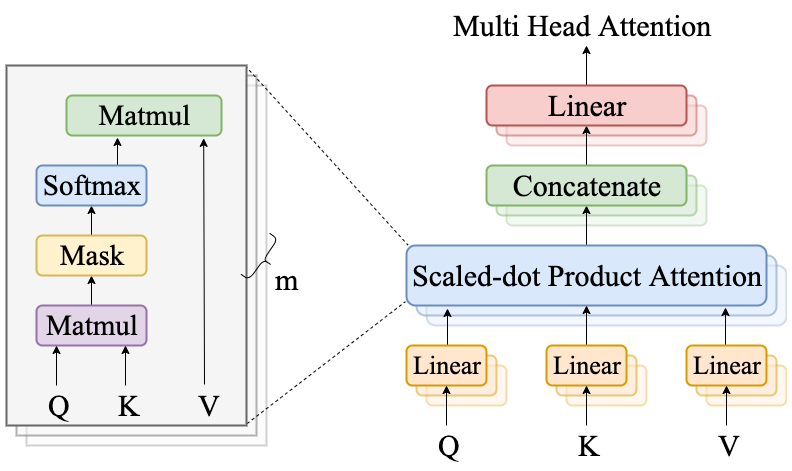}
    \caption{Internal structure of the Multi-Headed Self-Attention Mechanism in
a Transformer block.}
    \label{fig:attn_trans}
\end{figure}

\subsection{Transformers}

Previously, LSTMs were the only way to effectively capture the semnatic cues in various sentiment analysis tasks. However, \cite{vaswani2017attention} proposed a novel architecture Transformer, which replaced the complex recurrent computations with attention mechanism. Transformers play a pivotal role in our architecture and have the added advantage of attention-directed classification of the reviews. The attention function is a mapping from the $d$-dimensional input Values $V$ onto the output ($\mathcal{A}$ which is parameterised by a set of Key - Value ($K$, $V$) pairs. The Attention can essentially be defined as a weighted sum of Values $V = \{v_i | i \in \{1,2,...,d\}\}$ such that,

\begin{equation}
    Attention(K, Q, V) = \text{SOFTMAX}\left(\frac{QK^T}{\sqrt{d}}\right)V
\end{equation}

where $k, Q, V \in {\rm I\!R}^{d}$. Furthermore, in order to capture multiple perspectives, instead of averaging the attention scores, we use multi-head self-attention mechanism. Linear projections of Queries, Keys and Values are taken in order to incorporate the different perspectives of the $m$ attention heads which are then concatenated to form the final output. 

\begin{equation}
    MultiHead(Q,K,V) = \text{Concat}(\text{head}_1, \text{head}_2, ..., \text{head}_m )\mathbf{W}^O
\end{equation}
where $\text{head}_l = \mathcal{A}(Q\mathbf{W}_l^Q, K\mathbf{W}_l^K, V\mathbf{W}_l^V)$, such that the project matrices $\mathbf{W}_l^Q$, $\mathbf{W}_l^K$, $\mathbf{W}_l^V \in {\rm I\!R}^{\frac{d}{m} \times d}$, $\forall l \in \{1,2, ..., m\}$, and $\mathbf{W}^O \in {\rm I\!R}^{d \times d}$. The parallel computation of these heads is analogous to performing a single attention. Hence, allowing the model to capture positional and temporal information without increasing time complexity.

Due to the very long nature of reviews and the sparse suggestions present in them, the attention layers lead to significant enhancement in the performance of our classifier over the baselines tested. Attention-directed classification allows our model to filter out the necessary parts of the review to perform further analysis.

\begin{figure}[t]
    \centering
    \includegraphics[width=0.6\linewidth]{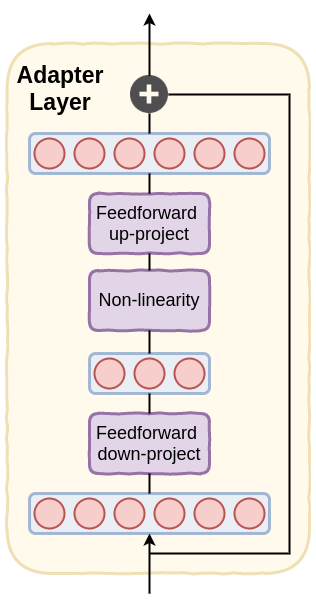}
    \caption{Internal structure of the Adapter layer used for Transfer Learning.}
    \label{fig:adapter}
\end{figure}

\subsection{Parameter-Efficient Transfer Learning}
Traditionally transfer learning in Natural Language Processing can be done via either of the two techniques, feature-based transfer and fine-tuning. Both these techniques require a new set of weights for each downstream task. Adapter-based transfer learning however allows us to overcome these drawbacks by providing a parameter efficient solution.

The internal structure of Adapter layer is shown in Figure \ref{fig:adapter}. Parameter tuning involves adding these adapter layers between an existing architecture and training its parameters. Mathematically, we can represent a neural network with a function $\phi_w(x)$, parametrised by $w$. Adapter tuning involves defining a new function $\psi_{w,v}(x)$, where the parameters $w$ are copied from the pre-trained network. The initial parameters $v_0$ are chosen such that they resemble the original function $\psi_{w, v_0} \approx \phi_w(x)$.

As shown in Figure \ref{fig:adapter}, the purple layers are fine-tuned during the training process. Internally, the adapter employs a bottleneck architecture. It projects the given $d$-dimesional input to a smaller $m$-dimensional input, applies a non-linearity and then projects it back to the original $d$ dimensions. Therefore, a total of $2md + d + m$ parameters are added per Adapter layer which is a significant improvement over training the original architecture end-to-end.

\begin{table}[t]
    \centering
    \begin{tabular}{c c}
        \toprule
        \toprule
        \bf Dataset Identifier & \bf Suggestion : Non-Suggestion \\
        \toprule
        Travel Train & 1314/3869 (0.34) \\
        Travel Test & 229/871 (0.26) \\
        \midrule
        Hotel Train & 448/7086 (0.06) \\
        Hotel Test & 404/3000 (0.13) \\
        \midrule
        Electronics Train & 324/3458 (0.09) \\
        Electronics Test & 101/1090 (0.09) \\
        \midrule
        Software Train & 1428/4296 (0.33) \\
        Software Test & 296/742 (0.39) \\
        \bottomrule
        \bottomrule
    \end{tabular}
    \caption{Details of the Open Domain Suggestion Mining (ODSM) dataset.}
    \label{tab:data}
\end{table}

\section{Data and Preprocessing}
To validate the proposed hypothesis we use the dataset curated by Negi at al. (b) \cite{odsm}. The dataset consists of reviews across four domains: Hotel, Electronics, Travel and Software annotated as Suggestion and Non-Suggestion. The details of the dataset can be found in Table \ref{tab:data}. Initial analysis reveals a huge imbalance in the data. This is due to the inherent deficit nature of suggestions on the various sources from where the reviews were collected. 

The following steps were involved in the preprocessing of the reviews before feeding them into the pipeline:

\begin{itemize}
    \item \textbf{Tokenization}: A tokenizer was used to parse the reviews into tokens. 
    \item \textbf{Spelling Correction}: Closer analysis of the text revealed a lot of spelling errors in the reviews. The spellings were corrected using a sequence matcher on the words and cross-referencing them to the words in the NLTK English corpus.
    \item \textbf{Lemmatization}: The tokenised tweets were then passed to the WordNet Lemmatizer to convert the words to their root forms.
    \item \textbf{Data Balancing}: Post the Lemmatization of the reviews, the imbalanced nature of the data is handled by the discourse marker-based minority oversampling as explained in phase 1 of the methodology.
    \item \textbf{Embedding Generation}: Word2vec model was then fine tuned on the balanced data to get custom embeddings for the reviews.
\end{itemize}

\begin{figure*}[t]
    \centering
    \includegraphics[width=\linewidth]{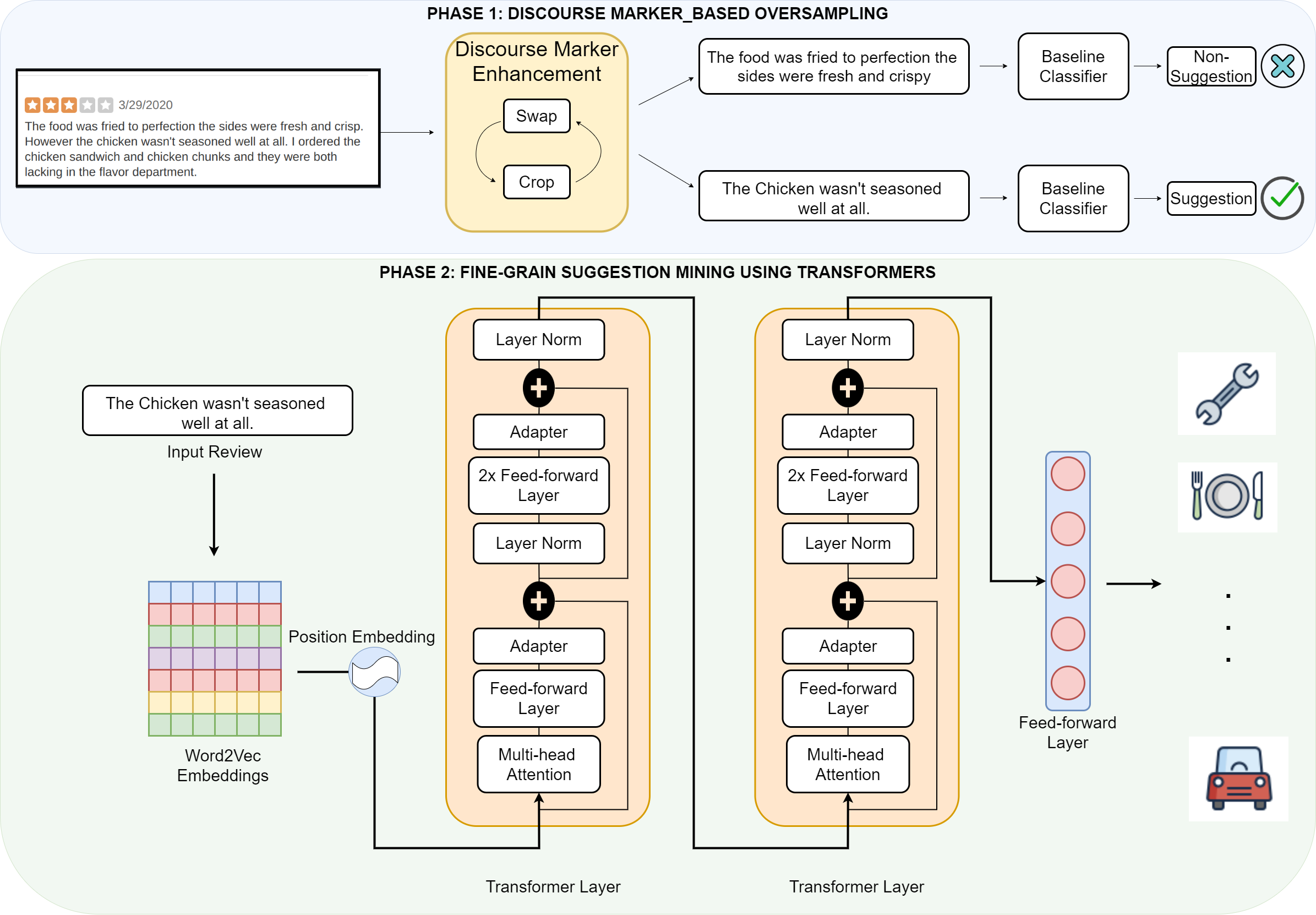}
    \caption{Overall architecture demonstrating the various phases of the proposed pipeline.}
    \label{fig:arch}
\end{figure*}

\section{Overall Pipeline}
\subsection{Phase 1: Discourse Marker based Over-Sampling}
The first phase of the pipeline involved handling the imbalanced nature of the dataset. This involved a three step process as described in Algorithm \ref{alg:data_aug}.The steps are as follows:
\begin{itemize}
    \item \textbf{Training Baseline Classifier}: The first step of phase 1 involves training a baseline sentiment analysis classifier. We use the architecture proposed by \cite{maitree} for the same.
    \item \textbf{Discourse Marker Enhancement}: For each review in the domain $d_i$, we check for the presence of traditional discourse markers like ``and", ``but", ``because". Once a discourse marker is found, the review is split into three parts $\{ s_{h_i}, m, s_{t_i} \}$ where $s_{h_i}$, and $s_{t_i}$ represent the head and tail discourse respectively. Once we obtain the head and tail discourse, we perform two operations namely \textbf{SWAP} and \textbf{CROP} on the segmented review. The SWAP operation involves swapping the head and tail discourse of the review. The CROP operation however involves cropping of the head and tail discourse from the compound review.
    \item \textbf{Inference-based Pruning}: Once the SWAP and CROP operations are applied on the review, they are pruned using the trained baseline classifier and the new sub-reviews classified as suggestion by $\mathcal{C}$ are added to the respective dataset.
\end{itemize}

This process is repeated for all the domains $d_i \in \mathcal{D}$ to obtain a balanced augmented dataset.

\begin{table*}[t]
    \centering
    \begin{tabular}{c c c c c c}
        \toprule
        \toprule
        \bf Method & \bf Hotel & \bf Electronics & \bf Travel & \bf Software & \bf Pooled (Fine-Grain) \\
        \toprule
        \bf Baseline (SVM) \cite{negi-buitelaar-2015-towards} & 0.79 & 0.78 & 0.66 & 0.72 & 0.33 \\
        \bf Baseline (LSTM) \cite{odsm} & 0.79 & 0.77 & 0.64 & 0.75 & 0.68 \\
        \bf FastText & 0.85 & 0.85 & 0.57 & 0.78 & 0.81 \\
        \bf FLAIR & 0.86 & 0.86 & 0.68 & 0.82 & 0.82 \\
        \bf Casual Transformer & 0.88 & 0.86 & \bf 0.80 & \bf 0.91 & 0.84 \\
        \bf SMOTE + FastText & 0.85 & 0.83 & 0.73 & 0.87 & 0.85 \\
        \bf Jain et al. \cite{maitree} & 0.86 & 0.83 & 0.71 & 0.88 & 0.85 \\
        \bf SMOTE + FLAIR & 0.87 & 0.84 & 0.76 & 0.89 & 0.87 \\
        \bf SMOTE + Casual Transformer & 0.89 & \bf 0.88 & 0.78 & 0.88 & 0.90 \\
        \bf Discourse Marker + FastText & 0.89 & 0.87 & 0.78 & 0.89 & 0.87 \\
        \bf Discourse Marker + FLAIR & \bf 0.91 & 0.87 & 0.77 & \bf 0.91 & 0.89 \\
        \bf Discourse Marker + Casual Transformer & \bf 0.91 & \bf 0.88 & \bf 0.80 & 0.90 & \bf 0.91 \\
        \bottomrule
        \bottomrule
    \end{tabular}
    \caption{Performance evaluation using F1 score.In all the cases, the discourse marker-based oversampling leads to a significant improvement over the baseline classifiers. Furthermore, qualitative analysis reveals the confusing nature of the reviews of Travel domain as the reason for the unusually low F1 values.}
    \label{tab:res}
\end{table*}

\begin{center}
\begin{algorithm}[ht]
\SetAlgoLined
 \textbf{Input:} Reviews ( $\mathcal{R}$ ) \\
    Pretrain a baseline suggestion classifier $\mathcal{C}$ \\
    Let $\mathcal{M}$ be the set of traditional Discourse Markers \\
    \For{$d$ in $\mathcal{D}$} 
    {
        \For{$r_i$ in $\mathcal{R}$} 
        {
            \For{$m$ in $\mathcal{M}$}
            {
                \uIf{$\mathcal{C}(r_i) = 1 \land d \in r_i$}{
                    $r_i = \{ s_{h_i}, m, s_{t_i} \}$ \\    
                    \tcp{SWAP Operation}
                    Add \{ $s_{t_i}, d, s_{h_i}$ \} to the dataset. \\
                    \uIf{$\mathcal{C}(s_{h_i}) = 1$}
                    {
                        \tcp{CROP Operation}
                        Add $s_{h_i}$ to the dataset. \\
                        
                    }
                    \uIf{$\mathcal{C}(s_{t_i}) = 1$}
                    {
                        \tcp{CROP Operation}
                        Add $s_{t_i}$ to the dataset. \\
                    }
                  }
            }
        }
    }
 \caption{Discourse Marker based Over-Sampling}
 \label{alg:data_aug}
\end{algorithm}
\end{center}

\subsection{Phase 2: Fine-grain Suggestion Mining using Transformers}
The second phase involves training the Transformer-based architecture on the augmented data. The initial step in this phase involves preserving the temporal nature of the input data. We use \textit{position embedding} ($\mathcal{P}$) for this purpose. $\mathcal{P} \in {\rm I\!R}^{ x \times d}$ is obtained by generating a sinusoidal position embedding with $1 \le i \le h$ positions for each vector-dimension $1 \le j \le d$:
\begin{equation}
    \mathcal{P}_{i,2j} = sin\left( \frac{i}{10000^{2j/d}}\right)
\end{equation}
\begin{equation}
    \mathcal{P}_{i,2j+1} = cos\left( \frac{i}{10000^{2j/d}}\right)
\end{equation}

To train the Transformer module, we include adapters with dimensionality $d=32$ and scale the learning rate by a factor of $10$ to train the newly added uninitialised parameters effectively. Gradients are accumulated over two steps to simulate larger batch sizes, which helps bring down the categorical cross-entropy loss faster.

\section{Experiments and Results}
The impact of our proposed pipeline and the oversampling strategies was tested by conducting several experiments and ablation studies on the dataset. The same test-train split was used as provided in the original dataset, the details of which can be found in Table \ref{tab:data}. Similar to the prior work including Negi et al. \cite{odsm} and Jain et al. \cite{maitree}, we use F1 score for the comparison of the various models experimented upon. The results of the various experiments performed are summarised in Table \ref{tab:res}.

\subsection{Baseline Classifiers}
The initial experiments involved replicating the baseline results on individual domains. These experiments were modelled as in-domain binary classification problems. Our Transformer-based architecture, used in phase 2 of our pipeine, was able to outperform all the other machine learning and deep learning based architectures including the ones used by Negi et al. \cite{odsm, negi-buitelaar-2015-towards}. We improved on the existing baseline architectures by significant margins across all the domains of suggestion mining. Jain et al. \cite{maitree} pointed out the confusing nature of suggestions in the Travel domain and the performance reduction it led to. However, our Transformer-based backbone was able to effectively mitigate this limitation.

\subsection{Fine-grained Analysis}
The experiments performed for in-domain suggestion mining were then extended towards open-domain suggestion mining via fine-grain analysis. In most of the cases, the results of the corresponding fine-grain analysis followed trends similar to those of in-domain analysis. However, in the case of simple classifiers like SVM and vanilla LSTM, the lack of sufficient parameters and inability to map complex decision boundaries. Approaches like FLAIR \cite{akbik-etal-2019-flair} and Transformer \cite{vaswani2017attention} were able to effectively learn the complex decision boundaries and yield significantly better results in the fine-grain analysis. However, the imbalance nature of the dataset was still driving the models away from optimal results. 

\begin{figure*}[t]
    \centering
    \includegraphics[width=\linewidth]{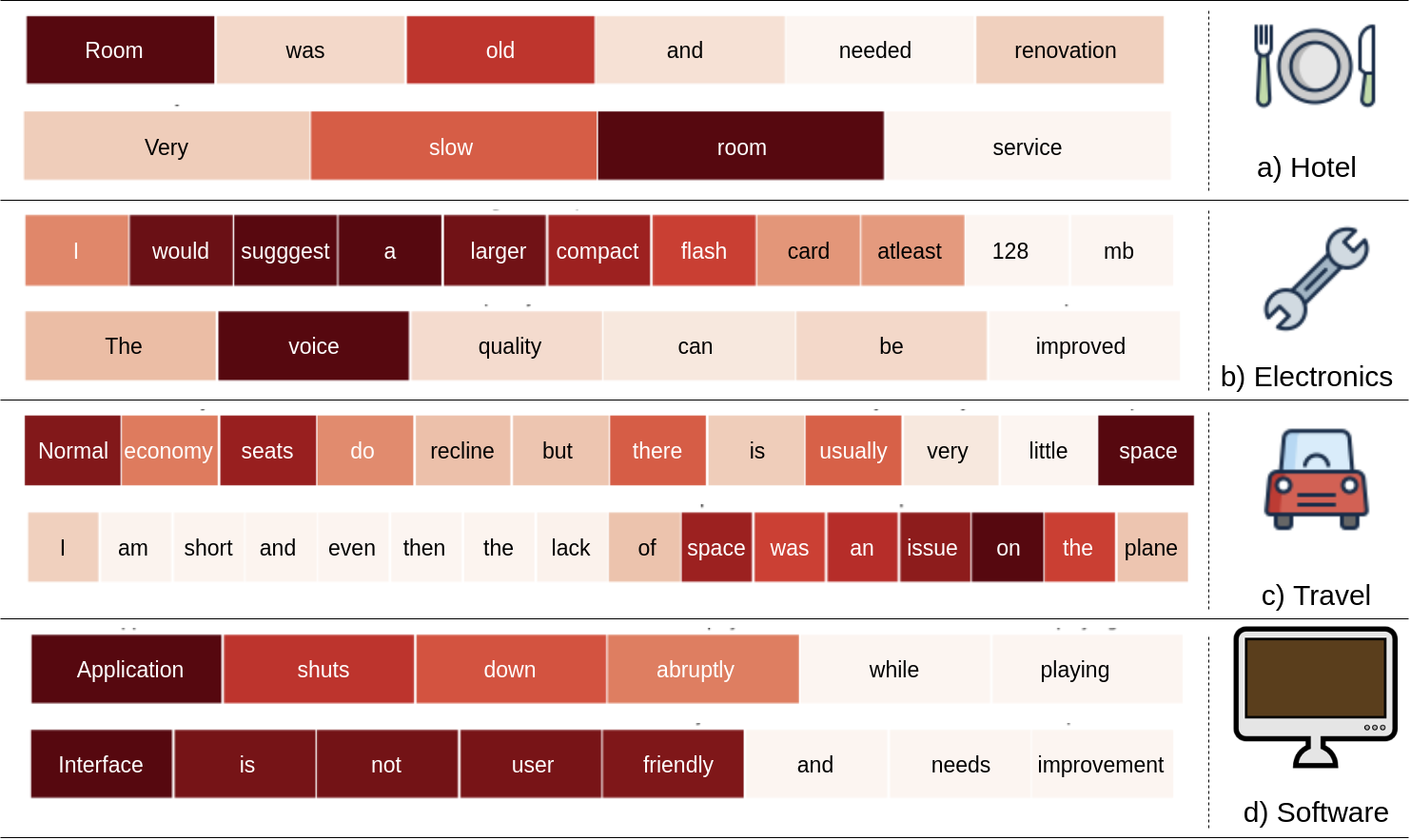}
    \caption{Qualitative Analysis of the word-wise heatmaps representing the relative attention scores for the suggestions in various domains under study.}
    \label{fig:attn}
\end{figure*}

\subsection{Oversampling Minority Classes}
Finally, the performance of our in-domain and multi-domain was enhanced using various data augmentation techniques. Minority class oversampling yielded a significant improvement in performance of our classifiers. Initially, SMOTE \cite{smote} was used as the standard minority oversampling technique. Though led to an improvement in the performance of our classifiers, we were unable to explain the enhanced results qualitatively due to the fact that SMOTE operates in euclidean space and is not interpretable. We then moved onto the Discourse marker based over-sampling approach explained in phase 1 of the pipeline. This approach turned out to perform comparably to SMOTE, without compromising model interpretability along the way.

\begin{table*}[ht]
    \centering
    \begin{tabular}{c c c c c c c c}
        \rowcolor{gray!50}
        
        \multicolumn{2}{c }{\textbf{Hotel}} & \multicolumn{2}{c }{\textbf{Electronics}} &  \multicolumn{2}{c }{\textbf{Travel}} &  \multicolumn{2}{c }{\textbf{Software}} \\

         \bf Token & {\bf SAGE} &  \bf Token & {\bf SAGE} &  \bf Token & {\bf SAGE} &  \bf Token & {\bf SAGE}  \\

         staff & 1.68 \DrawPercentageBar{0.7} & zen & 2.37 \DrawPercentageBar{0.9} & coach & 1.96 \DrawPercentageBar{0.8} & app & -2.95 \DrawPercentageBar{0.8} \\
         bathroom & 1.67 \DrawPercentageBar{0.69} & apex & 2.36 \DrawPercentageBar{0.88} & shorts & 1.94 \DrawPercentageBar{0.78} & api & 1.52 \DrawPercentageBar{0.66} \\
         breakfast & 1.65 \DrawPercentageBar{0.68} & g3 & 2.34 \DrawPercentageBar{0.87} & Egypt & 1.94 \DrawPercentageBar{0.78} & developers & 1.50 \DrawPercentageBar{0.65} \\
         lobby & 1.64 \DrawPercentageBar{0.65} & Canon & 2.34 \DrawPercentageBar{0.87} & Ireland & 1.93 \DrawPercentageBar{0.75} & buffering & 1.48 \DrawPercentageBar{0.63} \\
         renovated & 1.64 \DrawPercentageBar{0.65} & Nikon & 2.32 \DrawPercentageBar{0.85} & Prague & 1.91 \DrawPercentageBar{0.71} & Android & 1.46 \DrawPercentageBar{0.62} \\
         beds & 1.63 \DrawPercentageBar{0.62} & warranty & 2.29 \DrawPercentageBar{0.82} & customs & 1.89 \DrawPercentageBar{0.67} & iOS & 1.43 \DrawPercentageBar{0.58} \\
         decor & 1.62 \DrawPercentageBar{0.60} & DVD & 2.28 \DrawPercentageBar{0.78} & plane & 1.88 \DrawPercentageBar{0.65} & emulator & 1.42 \DrawPercentageBar{0.55} \\
         spatious & 1.60 \DrawPercentageBar{0.58} & viewfinder & 2.27 \DrawPercentageBar{0.75} & jacket & 1.87 \DrawPercentageBar{0.62} & feeds & 1.41 \DrawPercentageBar{0.52} \\
         buffet & 1.58 \DrawPercentageBar{0.55} & players & 2.25 \DrawPercentageBar{0.72} & dress & 1.87 \DrawPercentageBar{0.62} & browser & 1.40 \DrawPercentageBar{0.50} \\
         desk & 1.55 \DrawPercentageBar{0.50} & LCD & 2.23 \DrawPercentageBar{0.7} & tour & 1.85 \DrawPercentageBar{0.6} & notification & 1.36 \DrawPercentageBar{0.48} \\

    \end{tabular}
    \caption{Top 10 discriminating tokens used in the reviews of various domains extracted using along with their respective SAGE scores.}
    \label{tab:sage}
\end{table*}

\begin{figure*}[ht]
\centering  
\subfigure[Hotel]{\includegraphics[width=0.22\linewidth]{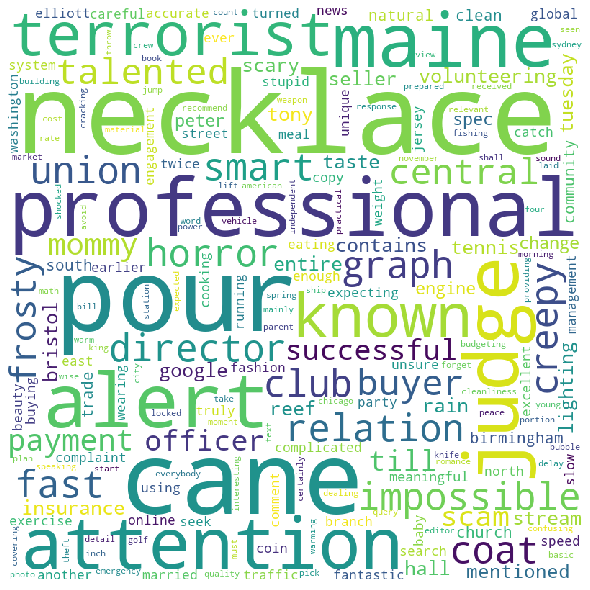}}
\subfigure[Electronics]{\includegraphics[width=0.22\linewidth]{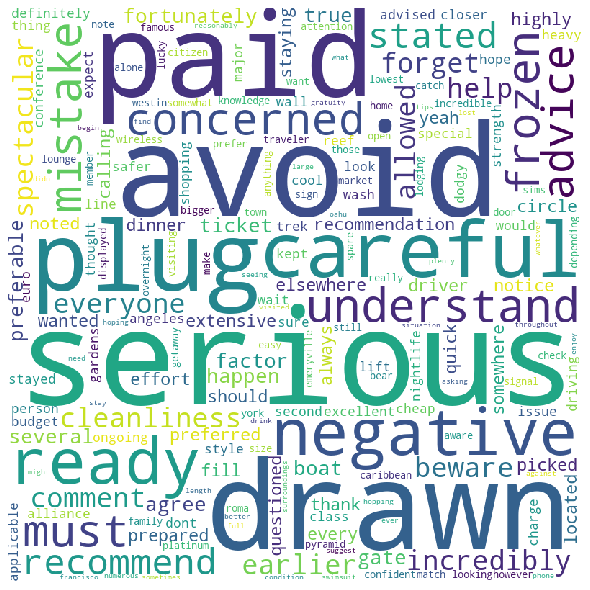}}
\subfigure[Travel]{\includegraphics[width=0.22\linewidth]{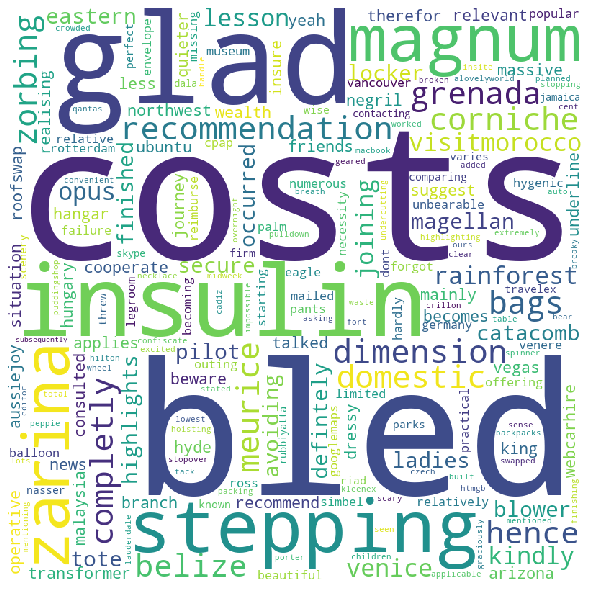}}
\subfigure[Software]{\includegraphics[width=0.22\linewidth]{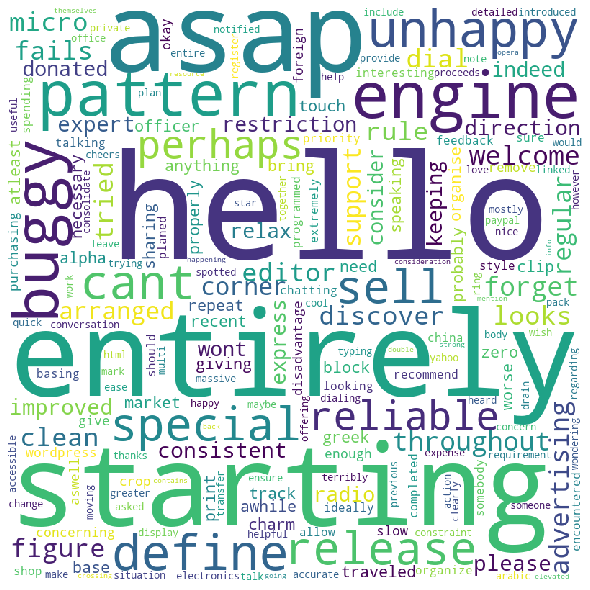}}
\caption{Attention-based word clouds for the various domains. The size of the words represent the relative attention values on the words as obtained from the multi-head self attention layer of the Transformer.}
\label{fig:wc}
\end{figure*}

\section{Discussion}

\subsection{Qualitative Analysis}
Due to the high dependence of our approach on the multi-head Attention mechanism in our pipeline, it was necessary to perform an extensive qualitative analysis of the Attention scores and the semantics they were capturing. The analysis involved analyzing the post-training attention inferences of complete reviews in the dataset, especially the very long reviews. Figure \ref{fig:attn} shows the sentence based analysis along with the corresponding heatmaps. It was seen that the list of words with the highest attention values had a high overlap with the words obtained via the SAGE analysis which provided construct validity to the usage of the attention-based transformer mechanism. The words and their relative attention scores were summarised in the form of domain-specific word clouds as sown
in Figure \ref{fig:wc}.

\subsection{SAGE Analysis}
We employ a Sparse Additive Generative Model (SAGE) to identify discriminating tokens between the suggestions across the various domains. The magnitude of the SAGE value of linguistic token signals the degree of its uniqueness, and in our case a higher SAGE value denotes representativeness of the word for the respective domain. Table \ref{tab:sage} reports the top 10 tokens for suggestions in each domain. Many of the discriminating keywords for each domain were the ones with the highest attention values found using the multi-head attention mechanism of the Transformer. The SAGE analysis therefore validified the attention modelling done by the Adapter-augmented Transformers.

\subsection{Limitations}
\begin{figure}[ht]
    \centering
    \includegraphics[width=0.9\linewidth]{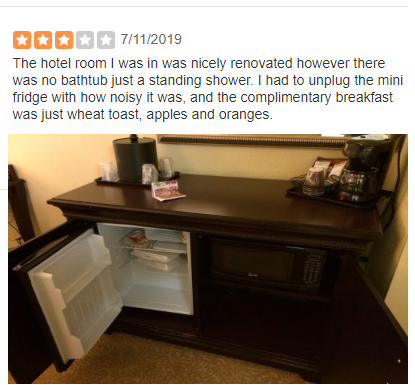}
    \caption{Review demostrating the limitation of our approach in capturing suggestions modelled as assertions.}
    \label{fig:limit}
\end{figure}

The primary limitation of our work lies in the inability of our proposed models to effectively capture suggestions modelled as assertions in very long reviews. As shown in Figure \ref{fig:limit}, cases where the suggestions in the form of assertions like ``I had to unplug fridge since it was noisy." are presented in a very long review tend to get mis-classified. This limitation can be tackled by taking into account multiple modalities while classification of the reviews. For instance in the above example, incorporating the accompanying picture of fridge along with the textual content will help improve the performance of the model significantly. This can be done by an architecture similar to the ones proposed in Agarwal et al. \cite{crisis, memis}.

\section{Conclusion and Future Work}
In this work, we propose a Transformer-based fine-grain analysis approach to tackle the task of open domain suggestion mining. We further mitigate the problem of imbalance dataset via a discourse marker based oversampling approach. Our experiments showed that our proposed pipeline outperforms the current state of the art by significant margins. We further perform extensive qualitative and quantitative experiments so as to prove the construct validity of our proposed pipeline. The future work involves incorporating reviewer profiling information into our classification models as proposed in works like Sinha et al. \cite{cikm}, Ghosh Chowdhury et al. \cite{hostile}. Furthermore, reviews are often associated to the geographical conditions and the corresponding issues with the products or services. Incorporating geographical information of the reviews as done in Gautam et al. \cite{gautam2019metooma} can thus help improve the performance of our model. Another possible direction can be using mixup techniques like the ones introduced in Jindal et al. \cite{DBLP:conf/aaai/JindalGSS20} instead of the discourse marker based technique to handle the data imbalance. This will lead to an increased adversarial robustness of the trained models thereby better downstream classification results.

\bibliographystyle{IEEEtran}
\bibliography{bare_conf}

\end{document}